\documentclass[a4paper,11pt,final]{article}

\usepackage{graphicx,amsmath,amssymb} 
\usepackage{epsfig}

\begin{document}

\title{Coevolutionary intransitivity in games: \\ A landscape analysis}
\author{Hendrik Richter\\
HTWK Leipzig University of Applied Sciences \\ Faculty of Electrical Engineering \&  Information Technology \\
Postfach 301166, D--04251 Leipzig, Germany\\
Email: richter@eit.htwk-leipzig.de}

\maketitle

\begin{abstract}
Intransitivity is supposed to be a main reason for deficits in coevolutionary progress and inheritable superiority.  Besides, coevolutionary dynamics is characterized by interactions yielding subjective fitness, but aiming at solutions that are superior with respect to an objective measurement. Such an approximation of objective fitness may be,  for instance, 
generalization performance. In the paper a link between 
 rating-- and ranking--based measures of intransitivity and fitness landscapes that can address the dichotomy  between subjective and objective fitness
is explored. The approach is illustrated by numerical experiments involving 
 a simple random game with  continuously tunable  degree of randomness. 

\end{abstract}

\section{Introduction}
Despite earlier promises and optimism, using coevolutionary algorithms (CEAs) for evolving candidate solutions towards an optimum remains a complicated and almost arcane matter with generally unclear prospects of success~\cite{mic09,popo10,vanwijn08}. This is prominently caused by a defining feature of coevolution. CEAs  are driven by fitness that originates from interaction of candidate solutions with other candidate solutions. In other words,  the fitness obtained from these interactions is subjective as it depends on which candidate solutions are actually interacting and when in coevolutionary run--time the interactions take place. The interactions can be understood as to constitute tests, which leads to labeling such kinds of coevolutionary problems as test--based problems~\cite{popo10,rich14a}. Test--based problems particularly occur in game playing or game--like contexts, for instance in situations where the players' strategies are subject to (competitive) coevolutionary optimum finding~\cite{popo10,samo13}.  It has been argued that in games with players and strategies the player space can be understood a phenotypic according to the framework of fitness landscapes~\cite{kall01,rich14aa}, while the strategy space is genotypic~\cite{ant09,now10}. This view is adopted in the following discussion.

  Notwithstanding that  coevolutionary dynamics is induced by  subjective fitness, the aim of using a CEA is identifying candidate solutions that are superior in a more general sense. Hence, in coevolution next to the  fitness resulting from (a limited number of) tests, a second notion of fitness  is helpful. Such a fitness generalizing subjective fitness  occurs in test--based problems in different forms. For games with players and strategies, there is usually no absolute quality measurement, or an absolute quality measurement would require to evaluate all possible test cases, which is computationally infeasible~\cite{chong08,chong12,samo13}. To circumvent this problem and enable experimental studies of the relationships between subjective fitness and absolute quality measurements, number games have been proposed, for instance minimal substrates~\cite{rich14b,wat01}. These artificial problem settings postulate an absolute quality, called
objective fitness~\cite{dejong07}. In this line of reasoning, 
coevolutionary dynamics can be understood as aiming at progress in objective fitness by proxy of subjective fitness.   Consequently, a main difficulty in designing CEAs stems from the question of how well subjective fitness represents objective fitness.  In analogy to the (postulated) objective fitness of number games, for games with players and strategies all general quality measurements of subjective fitness are interpretable as objective fitness. This implies that for game playing there may  only be an approximation of objective fitness and different approximations are possible, for instance different instances of generalization performance~\cite{chong08,chong12}. 
Put another way, this interpretation suggests that in game playing there are different degrees of objective fitness. 

Application examples of CEAs have frequently reported experiments showing mediocre performance, mostly attributed to coevolutionary  intransitivity~\cite{jong04,funes05,wat01}. Generally speaking, intransitivity occurs when superiority relations are cyclic. 
 Such cyclic superiority relations have consequences for coevolutionary dynamics as intransitivity may occurs across subsequent generations. In such a case, it may be  that all solutions at generation $k+1$ are better than at $k$, and that the same applies for $k+2$ with respect to $k+1$. This, however, does not imply that the solutions at $k+2$ are strictly better than at $k$. Cyclic superiority relations that occur across generations  are connoted as coevolutionary (dynamic) intransitivity. 
In this paper, the problem of coevolutionary intransitivity is linked to the dichotomy between subjective and objective fitness. This is done by combining a rating-- and ranking--based measuring  approach of intransitivity proposed by Samothrakis et al.~\cite{samo13}  with a framework of codynamic fitness landscapes recently suggested~\cite{rich14b}. Codynamic fitness landscapes enable to analyse the relationship between objective and subjective fitness for all possible solutions of the coevolutionary search process. The landscape approach proposed particularly explores how  coevolutionary intransitivity is related to  the objective--versus--subjective--fitness  issue. 
The remainder of the paper is structured as follows. In the next section, the concept of codynamic landscapes composed of objective and subjective fitness is briefly recalled. In Sec. 3, intransitivity is discussed. For the discussion, a simple random game is introduced, where the degree of randomness can be continuously tuned. It is shown that intransitivity can be characterized by different types of intransitivity measures. Numerical experiments with the simple random game are presented in Sec. 4, and Sec. 5 concludes the paper with a summary.     

\section{Coevolution, codynamic landscapes and number games}
 This section focuses attention on  an approach recently suggested~\cite{rich14b}  that is useful for understanding coevolutionary dynamics through codynamic fitness landscapes.  Such landscapes allow studying the relationship between objective and subjective fitness which, in turn, mainly determines   coevolutionary dynamics.  

We define objective fitness as the triple of search space $S$ with search space points $s \in S$, neighborhood structure $n(s)$ and fitness function $f_{obj}(s)$. The objective fitness landscape can be considered as to describe the optimization problem to be solved by the CEA. This problem solving is based on coevolutionary interactions between potential solutions which yields subjective fitness. Hence, subjective fitness can be viewed as the way the CEA perceives the problem posed by the objective fitness. From this, it appears to be sensible to assume that the subjective landscape  possesses the same search space and neighborhood structure, but has a fitness function $f_{sub}$ that more or less strongly deviates from the objective fitness $f_{obj}$. This can be seen as the subjective fitness  usually overestimating or underestimating objective fitness.  Moreover, for a coevolutionary run, the deviation between objective and subjective fitness is dynamic. In other word, the coevolutionary dynamics dynamically deforms the subjective fitness landscape.  In the following, the link between subjective and objective fitness is exemplified for  a number game, which is also called a coevolutionary minimal substrate~\cite{rich14b,wat01}.

In this
 number game a population $P$  of players is considered that inhabits the search spaces $S$. The search space is one--dimensional and real--valued. At each instance of the game  $k=0,1,2,\ldots$,   the players
of  $P(k)$ may have possible values $s \in S$.  An objective fitness function is defined over the search space, that is $f_{obj}(s)$, which consequently casts an objective fitness landscape. The subjective fitness of $P(k)$  is the result of  an interactive number game. Therefore,  for each calculation of the subjective fitness $f_{sub}(s)$ for a player from $P$, a sample  $\sigma(E)$ of evaluators from $E \subseteq P $ is randomly selected. This sample is statistically independent from the sample for the next calculation. Denote $\mu$ the size of the sample $\sigma(E)$ out of $\lambda$ evaluators, with $\mu\leq \lambda$. 
 The number game further defines that the fitness  $f_{sub}(s)$ with respect to the sample $\sigma(E)$ is calculated by counting the (averaged) number of members in $\sigma(E)$ that have a smaller objective fitness $f_{obj}(\sigma_i(E))$, $i=1,2,\ldots,\mu$, than the objective fitness $f_{obj}(s)$~\cite{rich14b}: 
\begin{equation}\label{eq:subfitnnumber} f_{sub}(s)=\frac{1}{\mu} \sum_{i=1}^{\mu} \text{eval}(s,\sigma_i(E))  \quad \text{with}\quad \text{eval}(s,\sigma_i)=\left\{ \begin{array}{ll}1 &\quad \text{if} \quad f_{obj}(s)>f_{obj}(\sigma_i) \\ 0 &\quad \text{otherwise}\end{array} \right. .\end{equation}
Note that the number game considered 
postulates objective fitness and defines by Eq. (\ref{eq:subfitnnumber}) how subjective fitness is obtained by a coevolutionary interaction.  In the next section, this perspective of subjective and objective fitness is applied to games with players and strategies. A population of players is engaged to evaluate their subjective fitness by interaction with other players. 
  
\section{Static and coevolutionary intransitivity in games}
A relation $R$ is called intransitive over a set $S$ if for three elements $\{s_1,s_2,s_3\} \in S$ the relation $(s_1Rs_2)\wedge (s_2Rs_3)$ does not always imply $s_1Rs_3$.  An instance of intransitivity  is superiority relations that are cyclic, which   
in its  most obvious and purest form appears in game playing.   Cyclic superiority relations here mean that for three players and three strategies  $s_1$, $s_2$ and $s_3$, the player using $s_1$ wins against $s_2$, and $s_2$ wins against $s_3$, but $s_1$ loses against $s_3$.   A simple example is a rock--paper--scissor game, where ``paper'' wins over ``rock'', ``scissor''  wins over ``paper'', but ``scissor'' loses against ``rock''. Thus, ``paper'', ``rock'' and ``scissor''  are  possible strategies a player can adopt in this game. Note that this kind of intransitivity is a feature of the preference in a single round of the game. Hence, such an intransitivity is static (and actually game--induced) and has no immediate link to (co-)evolutionary dynamics. Consequently, the next question is how these superiority relations resemble situations with coevolutionary intransitivity and can be understood by the dichotomy between subjective and objective fitness. To obtain  (co-)evolutionary dynamics, the players need to adjust their strategies, and the game needs to be played for more than one round.  In other words, studying iterated games is also interesting as it serves to juxtapose static (game--induced) intransitivity with coevolutionary (dynamic and search--induced) intransitivity.  

One way to build  a relationship between game results and fitness is to apply a rating system. Examples of  rating systems are the Elo system to evaluate chess players~\cite{elo78,lang12}, or the Bradley--Terry--Luce model of paired comparison~\cite{brad52,luce59}. Recently, it has been shown that this methodology is also useful for analyzing coevolutionary intransitivity~\cite{samo13}. A rating system creates a probabilistic model based on past game results that can be seen as a predictor of future results. Most significantly, the rating system also imposes a (temporal) ranking of the players. In the following, these ideas are applied to a simple random game where the degree of randomness can be tuned. The game consists of players using a strategy to perform against  all other players once, called a round robin tournament. The game outcome, which can be interpreted as payoff, is subject to the players' ratings and random. Given that the game has $N$ players, there are $\frac{N(N-1)}{2}$  games in a single round robin.  Define a percentage $p_{rand}$ of games that have a random result to obtain  $p_{rand}\cdot \frac{N(N-1)}{2}$ games whose outcome is chance with a predefined distribution. The remaining games end deterministically according to the rating difference between the players. Thus, such a game falls into the category of perfect and incomplete information. Viewed over a series of round robin tournaments, this interaction between rating--based determinism and random chance creates temporal ``rating triangles'', where a player scores high results (and has a high rating) over a certain time, but may also lose against a nominal weaker (low--rating) player, which over time may or may not show the same characteristics towards a third player. Such a behavior complies with coevolutionary (dynamic and actually search--induced) intransitivity.  In addition, the game can also reproduce rock--paper--scissor--like intransitivities. For $N$ players the maximal number of static intransitivities  is \begin{equation} \#_{intra \: max}= \left(\begin{array}{c} N \\ 3 \end{array} \right)=\frac{(N-2)(N-1)N}{6}, \end{equation} see~\cite{frank82,samo13}. Any three players form a triangle of cyclic superiority relations if they each win one game against the other two. Thus, for $N$ not very large, the (average) number of actual static intransitivities  can be determined by enumeration and gives a static intransitivity measure called the intransitivity index (itx)~\cite{frank82,samo13}. As an alternative, Samothrakis et al.~\cite{samo13} suggested to use a difference measure based on Kullback--Leibler divergence (kld) between the prediction made by the rating system and the actual outcome to measure static intransitivity.  Both quantities itx and kld are subjects of the numerical experiments reported in the next section.

 \begin{table}[tb]
\caption{Results of the simple random game for three instance of the game}
\label{tab:1rich}       
%
%
\begin{tabular}{|c|c||ccccc|c|c||ccccc|c|c|c|c||c|}
$\#$  & rt(0)& $\#$1 & $\#$2 & $\#$ 3 & $\#$4 & $\#$5 & sc(0) &   rt(1) & $\#$1 & $\#$2 & $\#$ 3 & $\#$4 & $\#$5 & sc(1) &   rt(2) & gp& rank & sc(2)\\ \hline
1 &  1600& x & 1 & 1& 1& 1& 4 & 1630 & x & 0 & 1 & 1&1&3& 1642 & 3.5& 1& 1\\
2 & 1600 &0 &x&1&0&1& 2 & 1600& 1 & x& 1 & 1&  0&3&1615& 2.5& 2& 4\\
3 &1600 &0 &0&x&1&1& 2& 1600& 0 & 0 & x& 0&0&0& 1570&1& 4.5& 0\\
4 &1600&0 &1&0&x&1& 2 & 1600& 0&0&1&x&1&2& 1600&2& 3& 2\\
5 &1600&0&0&0&0&x & 0 & 1570& 0 & 1&1&0&x&2& 1573&1& 4.5& 3\\
\hline

\end{tabular}
\end{table}
 A  game in a coevolutionary setting involves finding the strategy a player should adopt to score best according to a given understanding of performance. This clearly implies that the performance measurement should generalize a single round robin tournament. Thus, if there are several instances of round robin tournaments,  the overall results can also be accounted for by generalization performance~\cite{chong08,chong12}.  Each instance of a round robin can be scaled to a generation of coevolutionary run--time. Generalization performance is defined as mean score of a solution in all possible test cases. Because considering all possible test cases may be computationally infeasible, Chong et al.~\cite{chong08,chong12} used a statistical approach involving confidence bounds to estimate the amount of needed test cases for a given error margin. Given this understanding, and assuming that all strategies are equally likely to be selected as test strategies, the generalization performance  of strategy $i$ is: \begin{equation} \text{gp}_i=\frac{1}{\mathcal{K}}\sum_{k=1}^\mathcal{K} \text{sc}_i(k), \label{eq:gp}\end{equation}
where $\text{sc}_i(k)$ is the score the  $i$-th strategy yields in the $k$--th instance of a round robin tournament. The needed number of instances $\mathcal{K}$ depends on the bounds given by Chong et al.~\cite{chong08,chong12}. Generalization performance also builds a relationship between actual game results and fitness and therefore can be seen as an alternative to a rating system. 

 As an example of the simple random game
assume that there are five players ($N=5$), denoted as $\#1$ to $\#5$, which each act upon a unknown, but internally adjusting strategy. Further assume  that as an a--priori  evaluation all players are considered equal and have the same rating, say $\text{rt(0)}=1600$. These players are now engaged in a round robin tournament, where a win scores $1$ and a  loss  counts $0$.  The results achieved depend on the pre--game rating and random. Assume that these results were scored, see Tab. \ref{tab:1rich}, column 3--7. While player $\#1$ wins all games and $\#5$ loses all games, the players $\#2$--$\#4$ build a rock--paper--scissor triangle of cyclic superiority relations. These results somehow violate the expectations established by the initial (pre--game) rating, which could have been met by all players winning 2 out of 4 games. Furthermore, the results show clearly that the game cannot be completely deterministic with respect to the a--priori evaluation.  In other words, a rating and ranking approach subsumes the game history and can only be a predictor of future game results. The quality of prediction depends on the percentage of random game results. Also, from the results of the round robin, it is not evident whether player $\#1$ was as successful as it was because its strategy was (objectively) good, or because the strategies of the other players were (objectively) poor. All, the round robin gives is a comparison.

The a--priori rating were indicating that all players have the same rank (and hence there is no ranking difference between them), but the results were showing otherwise. Hence, the round robin tournament updates the rating, producing an after--game rating. According to the Elo system~\cite{elo78,lang12}, which is adopted here, this is done via first calculating the expected outcome
\begin{equation} \text{ex}_i(k)= \sum_j \frac{1}{1+10^{(\text{rt}_j(k)-\text{rt}_i(k))/400}}. \end{equation}
The quantity $\text{ex}_i(k)$ summarizes winning probabilities of  player $\#i$ with respect to all other players in round $k$.
For a single game, that is $j=1$, the quantity $\text{ex}_i$ is the expected winning probability for the player $\#i$ winning against player $\#j$. For the example with all players having the same rating, the expected outcome is also the same, namely $\text{ex}_i=2$.   The rating of the players is updated according to the difference between expectation and actual score:
\begin{equation}
\text{rt}_i(k+1)=\text{rt}_i(k)+K(\text{sc}_i(k)-\text{ex}_i(k)). \label{eq:elo}
\end{equation}
The $K$ is called the $K$--factor, which tunes the sensitivity between the rating and the results of a single round robin tournament.  Using $K=15$, the new rating $\text{rt}(1)$ gives differences between the players, see Tab. \ref{tab:1rich}, column 9. The best player $\# 1$ is now ranked highest, the poorest player $\#5$ is ranked lowest. Also note that the players $\#2$--$\#4$ engaged in the game--induced intransitivity triangle still share the same rating as before. Now assume the next round of the game. The strategies of the players have been adjusted, supposedly by a (competitive) coevolutionary search process. See the results in Tab.   \ref{tab:1rich}, column 10--14. The outcome generally confirms the impression of the first round with player $\#1$ still being strong. Violating the expectations are the good scores of the players $\#2$ and $\#5$, and the poor score of player $\#3$. Note that in this round the players  $\#1$, $\#2$, and $\#5$ as well as the players $\#2$, $\#4$, and $\#5$ form a rock--paper--scissor triangle of game--induced intransitivity. Contrary to the first round, these players now neither have the same score, nor the same pre--game or after--game rating. This after--game rating $\text{rt}(2)$ is calculated according to Eq. (\ref{eq:elo}) and given in Tab. \ref{tab:1rich}, column 16. 
\begin{figure}[tb]
\includegraphics[trim = 23mm 15mm 30mm 15mm,clip, width=6.0cm, height=5.8cm]{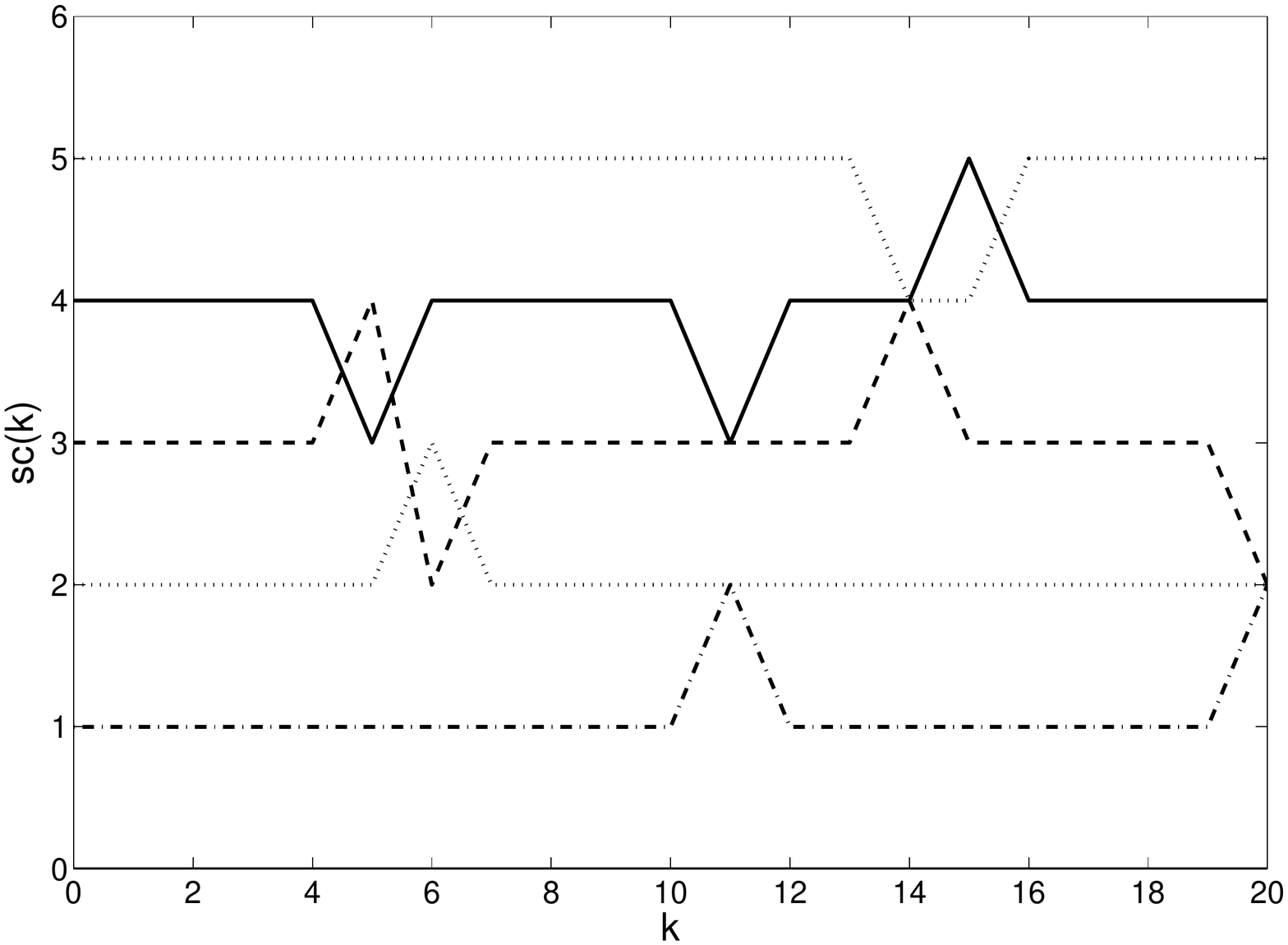}
\includegraphics[trim = 23mm 15mm 30mm 15mm,clip, width=6.0cm,height=5.8cm]{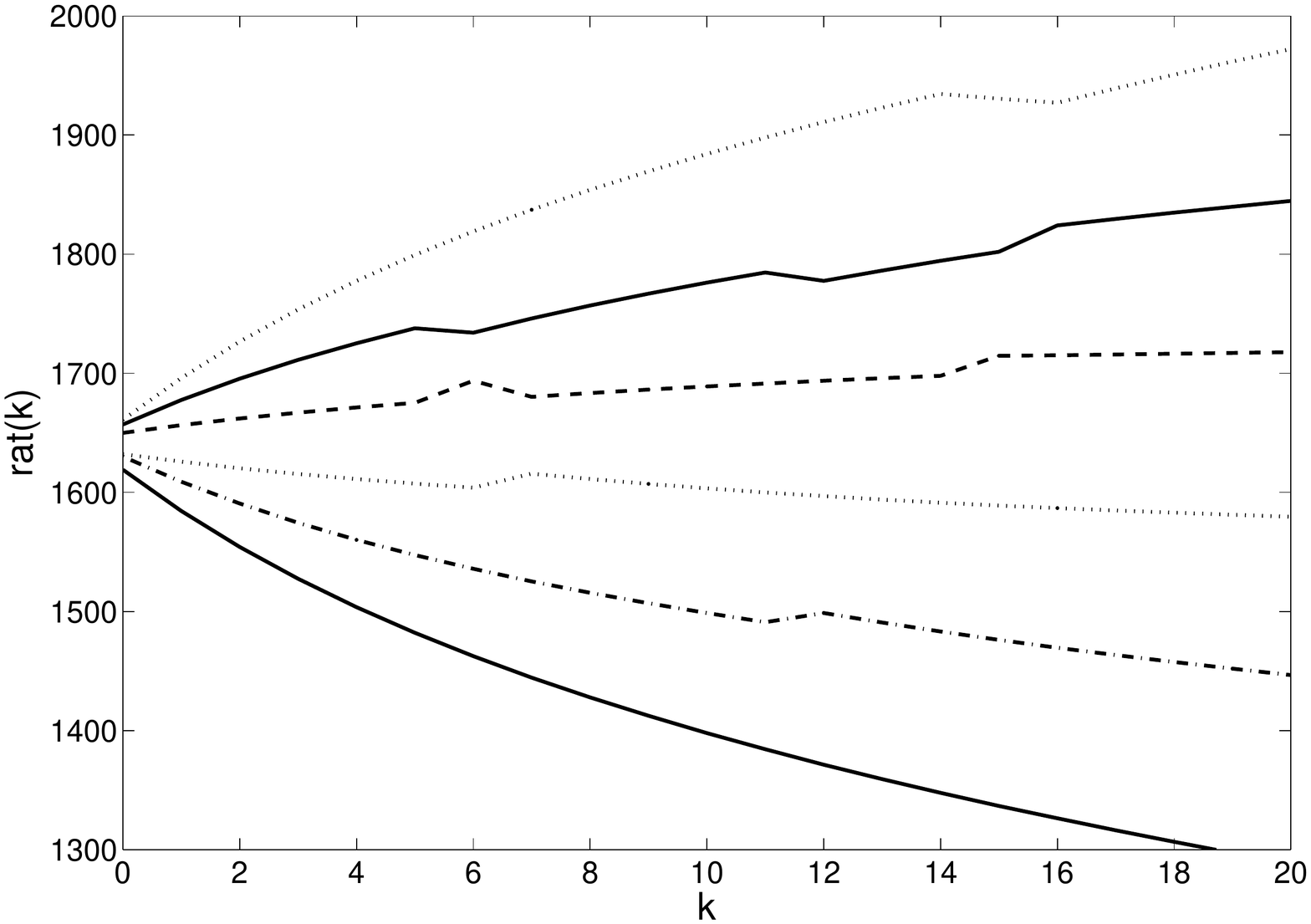}
\includegraphics[trim = 23mm 15mm 30mm 15mm,clip, width=6.0cm, height=5.8cm]{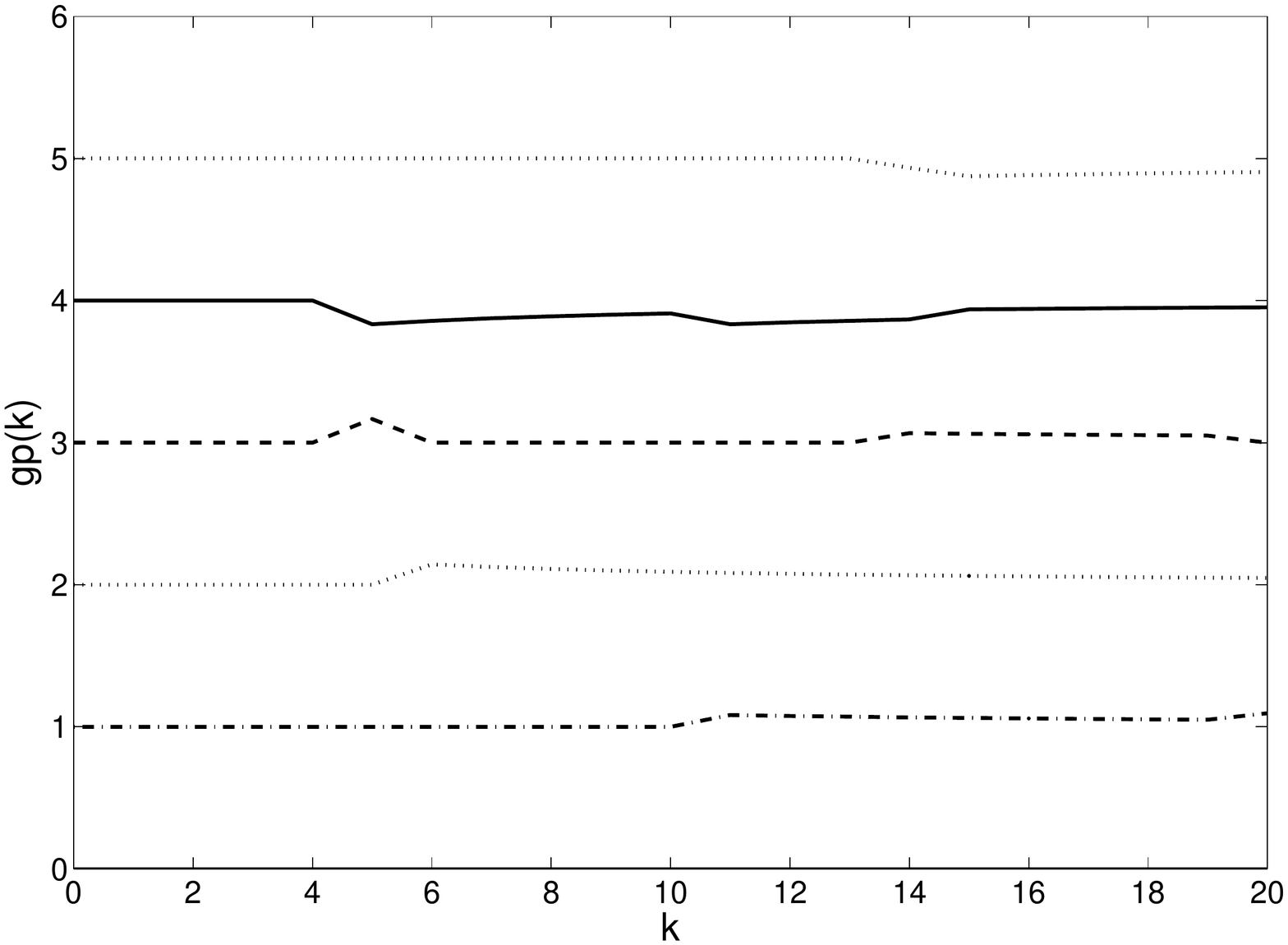}

\hspace{1.8cm}  (a) \hspace{3.6cm}  (b) \hspace{3.4cm}  (c)

\includegraphics[trim = 23mm 15mm 30mm 15mm,clip, width=6cm, height=5.8cm]{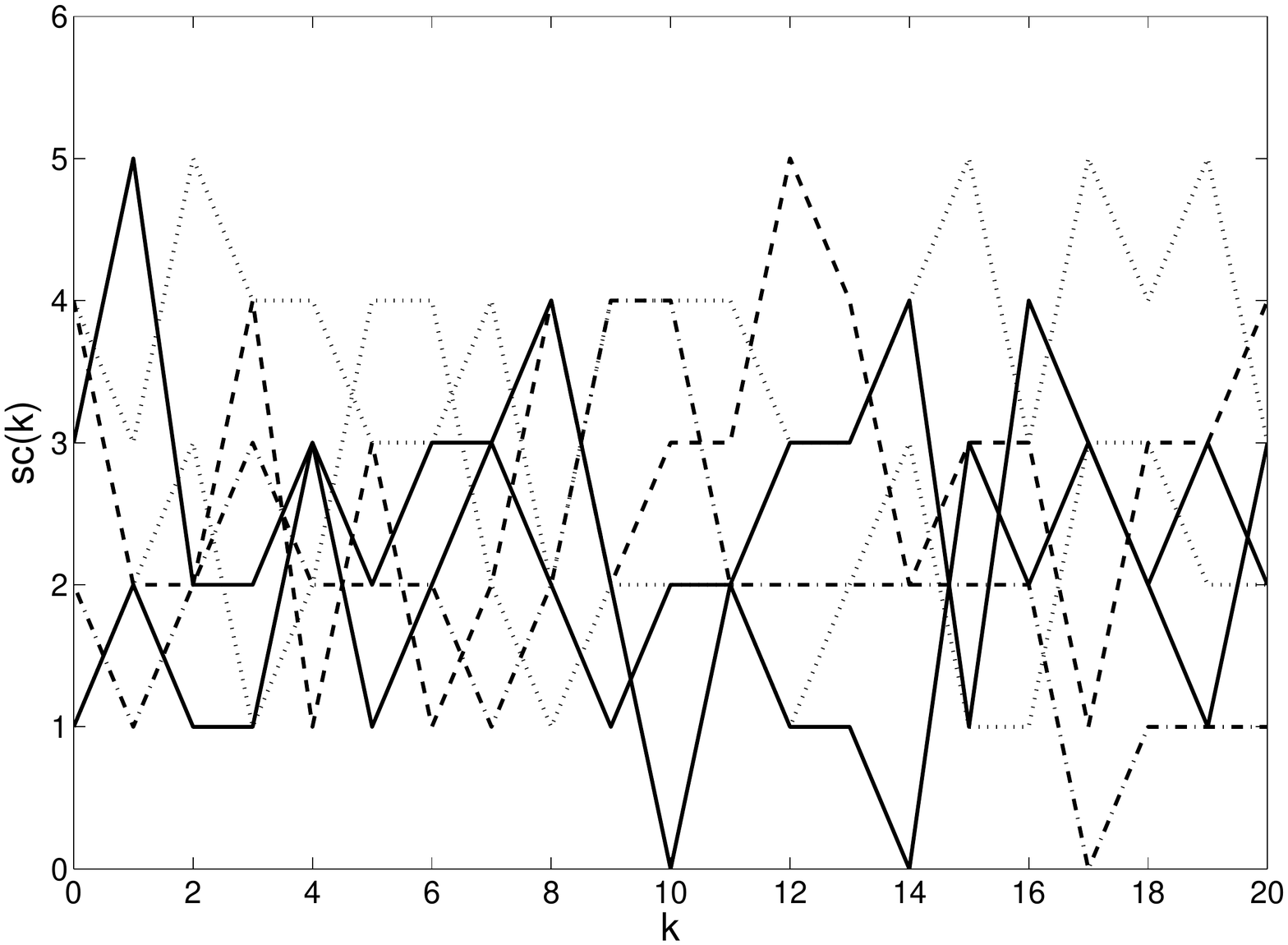}
\includegraphics[trim = 23mm 15mm 30mm 15mm,clip, width=6cm,height=5.8cm]{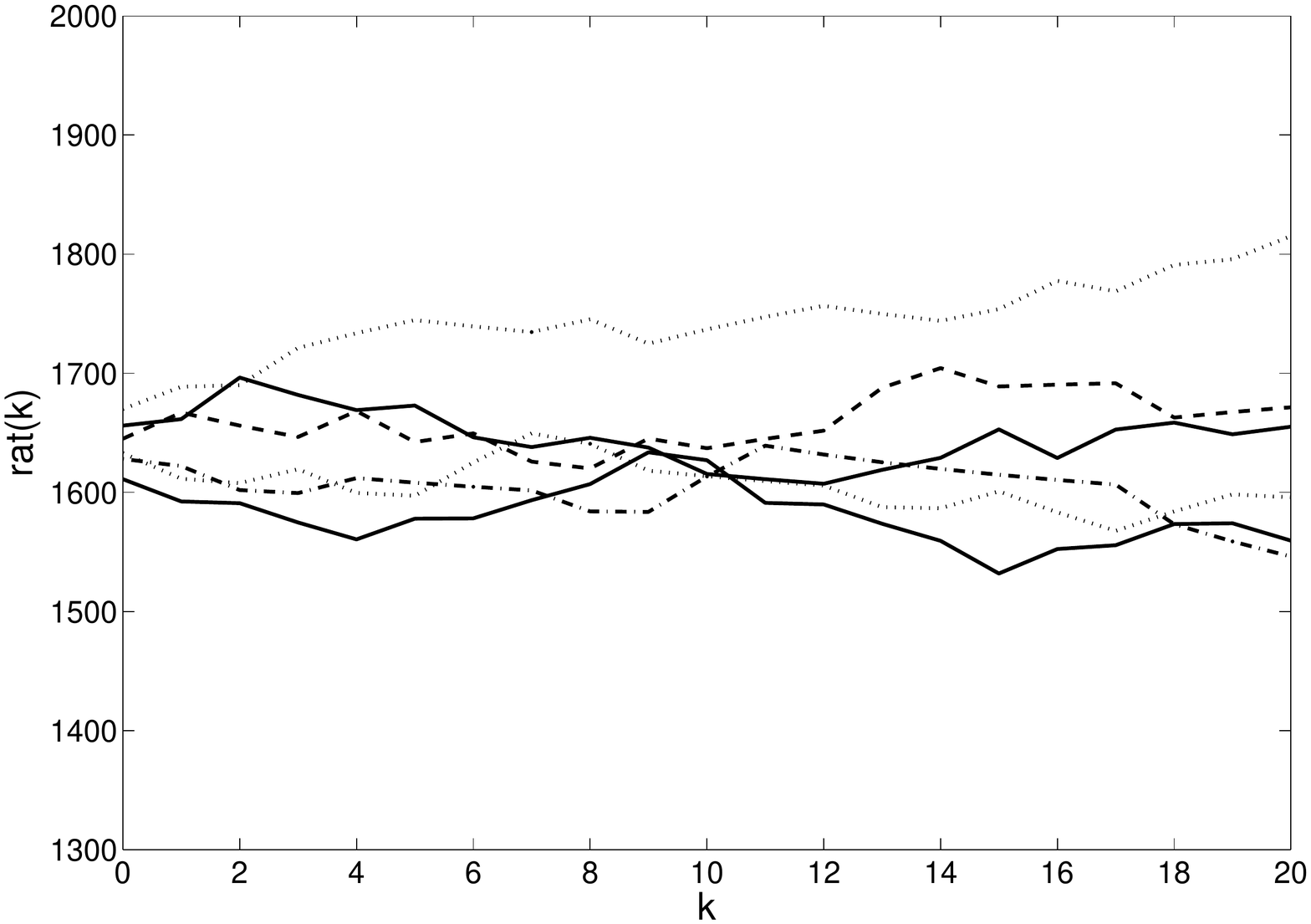}
\includegraphics[trim = 23mm 15mm 30mm 15mm,clip, width=6cm, height=5.8cm]{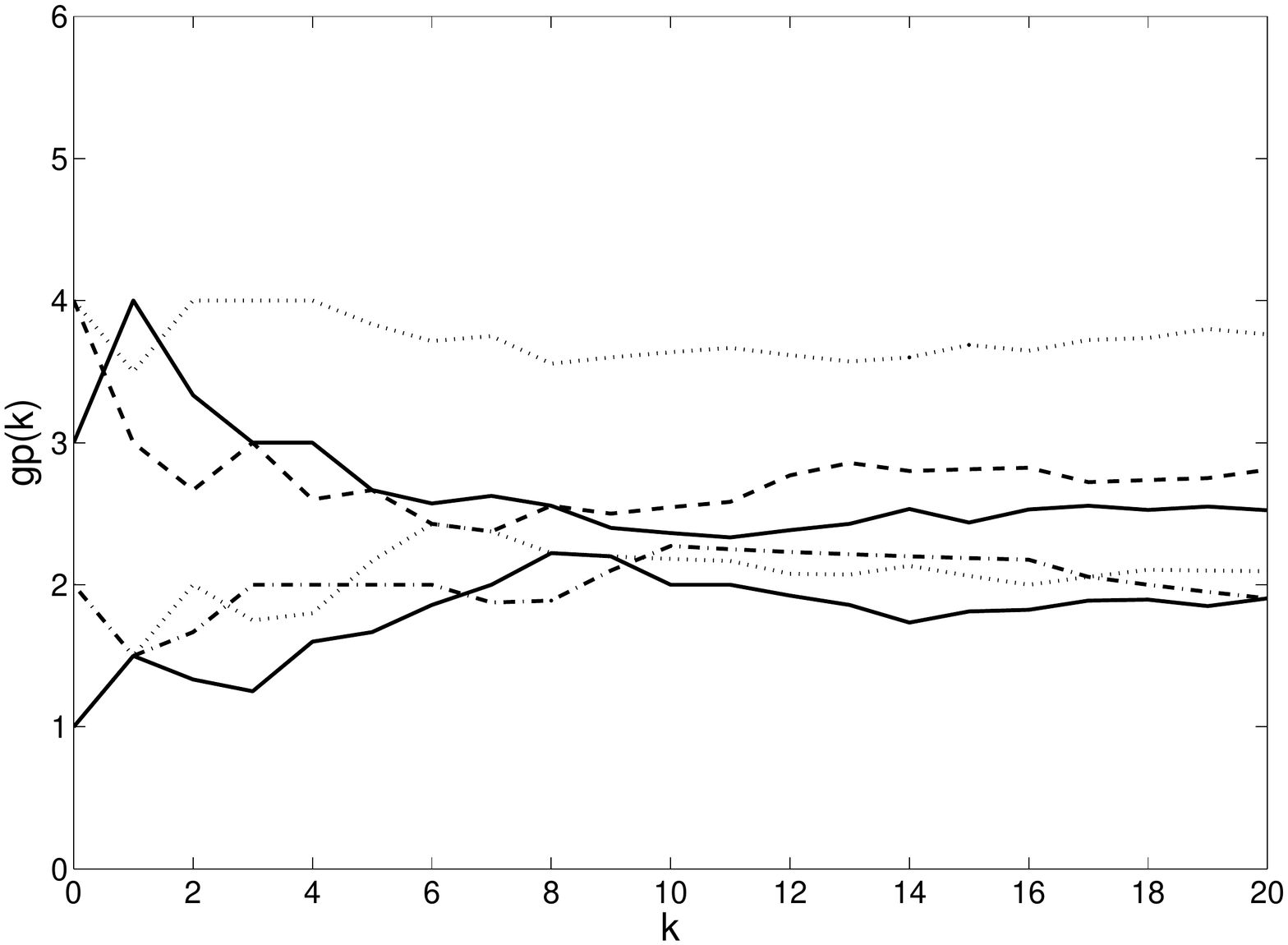}

\hspace{1.8cm}  (d) \hspace{3.6cm}  (e) \hspace{3.4cm}  (f)

\vspace{0.25cm}
 {\bf Fig. 1.} Scores,  ratings, and generalization performance for two percentages of random games: (a)--(c), $p_{rand}=0.01$; (d)--(f), $p_{rand}= 0.75$.
\end{figure}

As there are now two instances of the round robin tournament, an first estimation of the generalized performance $\text{gp}$ can be obtained by averaging  the scores according to Eq. (\ref{eq:gp}). 
The results are given in  Tab. \ref{tab:1rich}, column 17. Conforming with this account of quality evaluation, player $\#1$ is best, and can be ranked first, followed by players  $\#2$ and  $\#4$. The ranking with respect to generalization performance is given in  Tab. \ref{tab:1rich}, column 18. Note that the ranking gives average ranks for tied ranks as for player $\#3$ and $\#5$ ($\text{gp}=1$ leads to $\text{rank}=4.5$), which preserves the sum over all ranks. Further note that this ranking is almost equal to the ranking according to ratings, with the exception of players  $\#3$ and $\#5$ which have a very similar but not equal rating. 

Now, the simple random game is interpreted according to the landscape view of subjective and objective fitness.
Recall that subjective fitness is associated with fitness gained by individuals through interaction with others. According to this view, a round robin tournament yields subjective fitness: $f_{sub}=\text{sc(k)}$. Objective fitness, in turn, generalizes subjective fitness in terms of an absolute quality measurement. Possible candidates are the rating $f_{obj}=\text{rt(k)}$ or generalization performance $f_{obj}=\text{gp(k)}$. 
Defining subjective and objective fitness in such a way also gives raise to reformulating coevolutionary intransitivity. Generally speaking, coevolutionary intransitivity involves cycling of (objective) solution quality. This cycling may be caused by subjective fitness not adequately representing objective fitness. Hence, subjective fitness may drive evolution into search space regions visited before but evaluated differently, or generally into directions not favorable. Hence, coevolutionary (dynamic) intransitivity can be understood as temporal mismatches in order between subjective and objective fitness. Consider again the example of the game whose results are given in Tab. 1. Suppose  another instance of the round robin is played, and (omitting the specific results) the scores are in Tab. 1, column 19.  The rating $\text{rt(k)}$ is considered to be objective, while $\text{sc(k)}$ is subjective. For player $\#1$, the rating is $\text{rt}_1=1600 \leq 1630 \leq 1642$, while its score is $\text{sc}_1=4 > 3 > 1$, which is a temporal mismatch between objective and subjective fitness. If we were to suppose for a moment that the rating declared as objective fitness is indeed the quantity to achieve in coevolutionary search, and if a CEA were to use score for guiding this search, then player $\#1$ would  likely be misguided. On the other hand, for player  $\#2$, the rating $\text{rt}_2=1600 \leq 1630 \leq 1642$, and the score $\text{sc}_2=2 < 3 < 4$ show a match between subjective and objective fitness. These conditions can be reformulated employing a ranking function with tied ranks and gives a measure of coevolutionary intransitivity.  Hence, the player--wise temporal mismatch (ptm) can be defined as  the average number of rank mismatches:  $\text{rank}(f_{obj}(k), f_{obj}(k+1)), f_{obj}(k+2)) \neq \text{rank}(f_{sub}(k), f_{sub}(k+1)), f_{sub}(k+2))$ for each player for a given number of instances of round robins. 

 An alternative measure of coevolutionary intransitivity that is related to the ptm just discussed stems from the fact that coevolutionary selection is based on comparing subjective fitness values. Hence, the fitness ranking within one instance of the game (or one generation) gives indication as to what direction is preferred.  If difficulties in the search process are caused by how well subjective fitness represents objective fitness, then the difference in the ranking according to subjective fitness and the ranking according to objective fitness for each instance is also a suitable measure of coevolutionary intransitivity. Therefore, the quantity $|\text{rank}(f_{sub}(k))-\text{rank}(f_{obj}(k))|$ over all players for each instance $k$ is another measure of coevolutionary intransitivity  and is called  collective ranking difference (crd).
The observations discussed so far suggest some relationships, namely that game--induced, static intransitivity has ambiguous effect on coevolutionary pro\-gress, and that coevolutionary, dynamic intransitivity can be expressed as ranking differences between objective and subjective fitness. In other words, the quantities ptm and crd 
may be useful as measures of coevolutionary intransitivity.  All these relationships can be studied by numerical experiments, which are the topic of next section.

\begin{figure}[tb]
\includegraphics[trim = 23mm 15mm 30mm 15mm,clip, width=6.0cm,height=5.8cm]{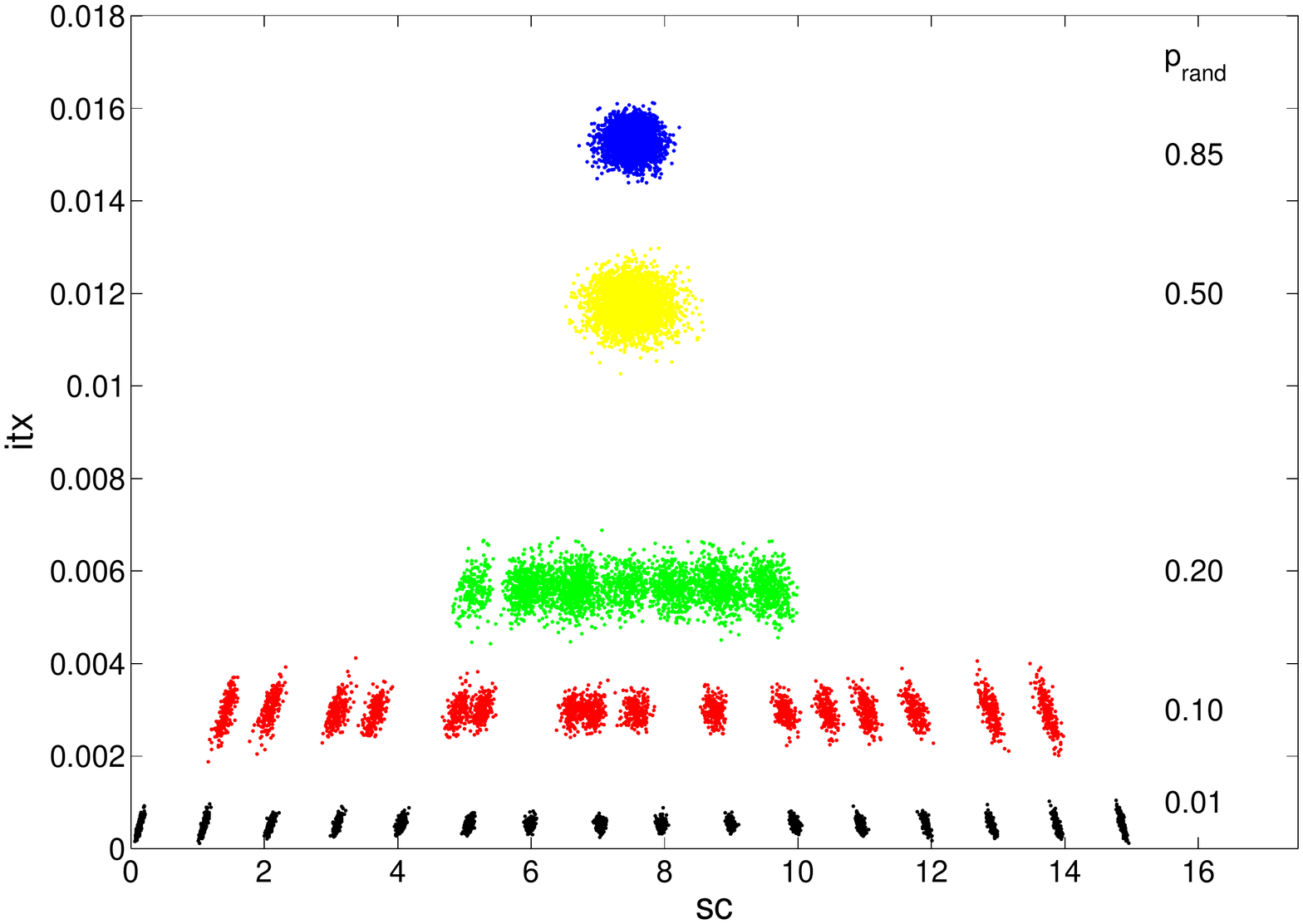}
\includegraphics[trim = 23mm 15mm 30mm 15mm,clip, width=6.0cm, height=5.8cm]{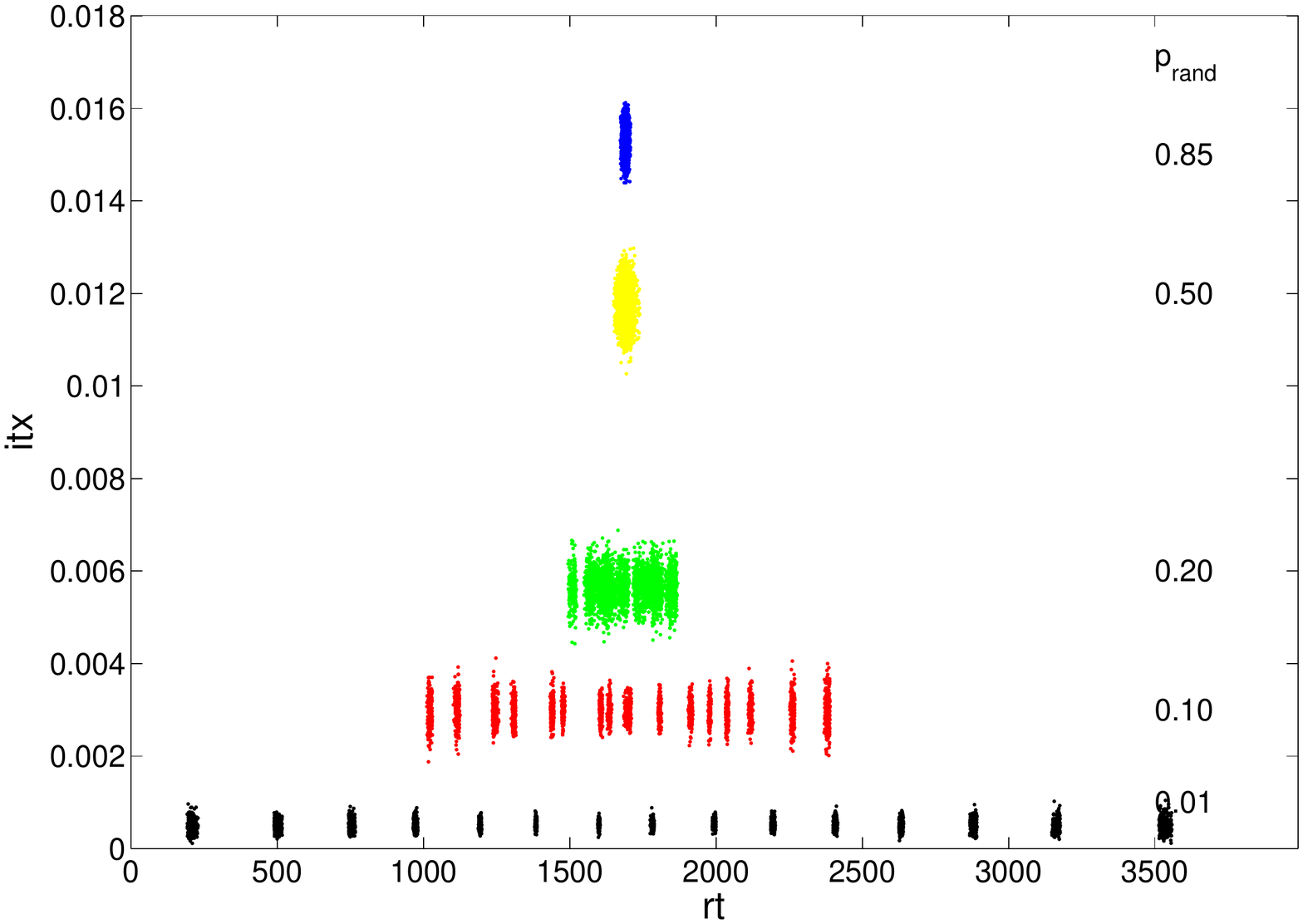}
\includegraphics[trim = 23mm 15mm 30mm 15mm,clip, width=6.0cm, height=5.8cm]{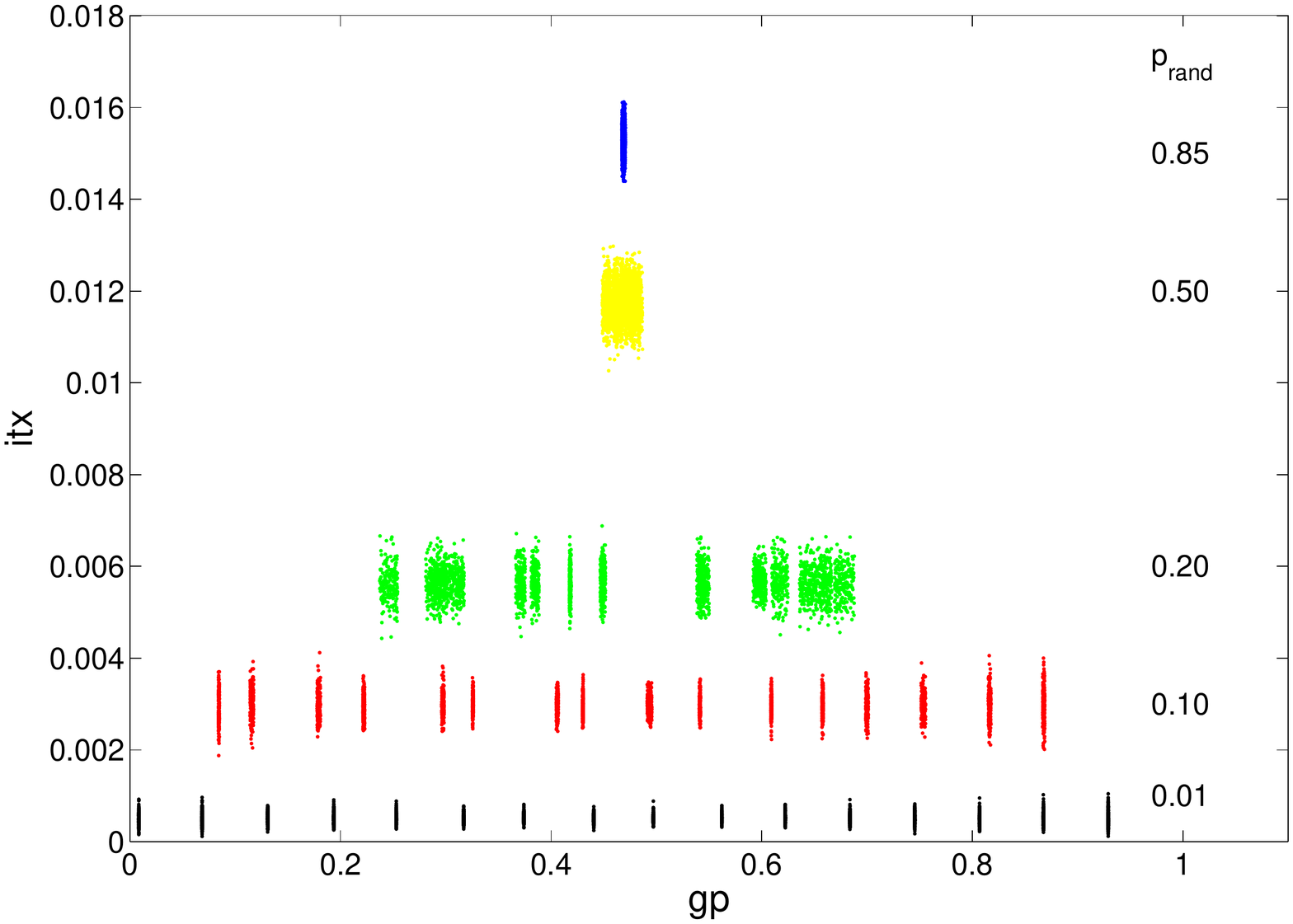}

\hspace{1.8cm}  (a) \hspace{3.6cm}  (b) \hspace{3.4cm}  (c)

\includegraphics[trim = 23mm 15mm 30mm 15mm,clip, width=6.0cm,height=5.8cm]{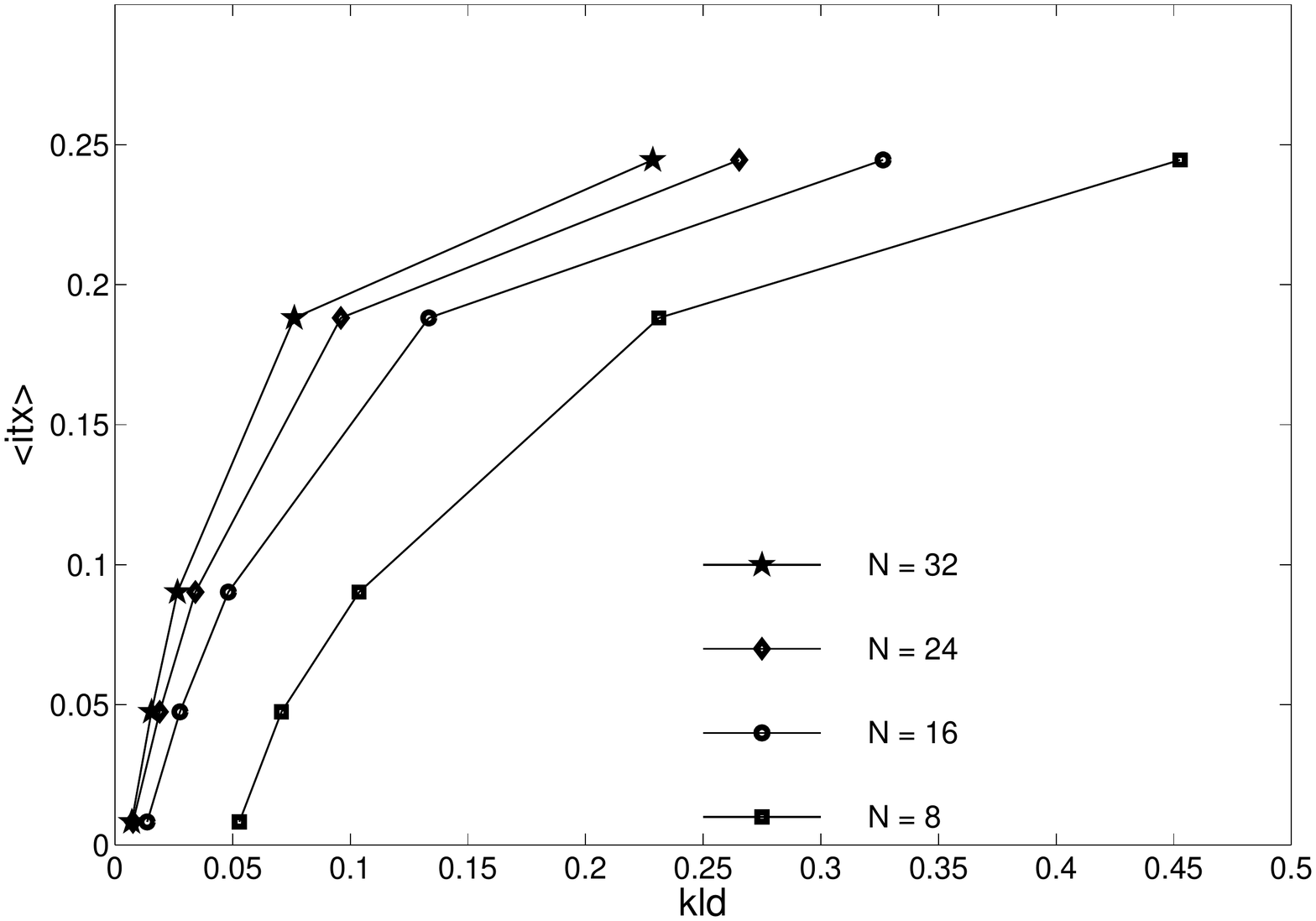}
\includegraphics[trim = 23mm 15mm 30mm 15mm,clip, width=6.0cm, height=5.8cm]{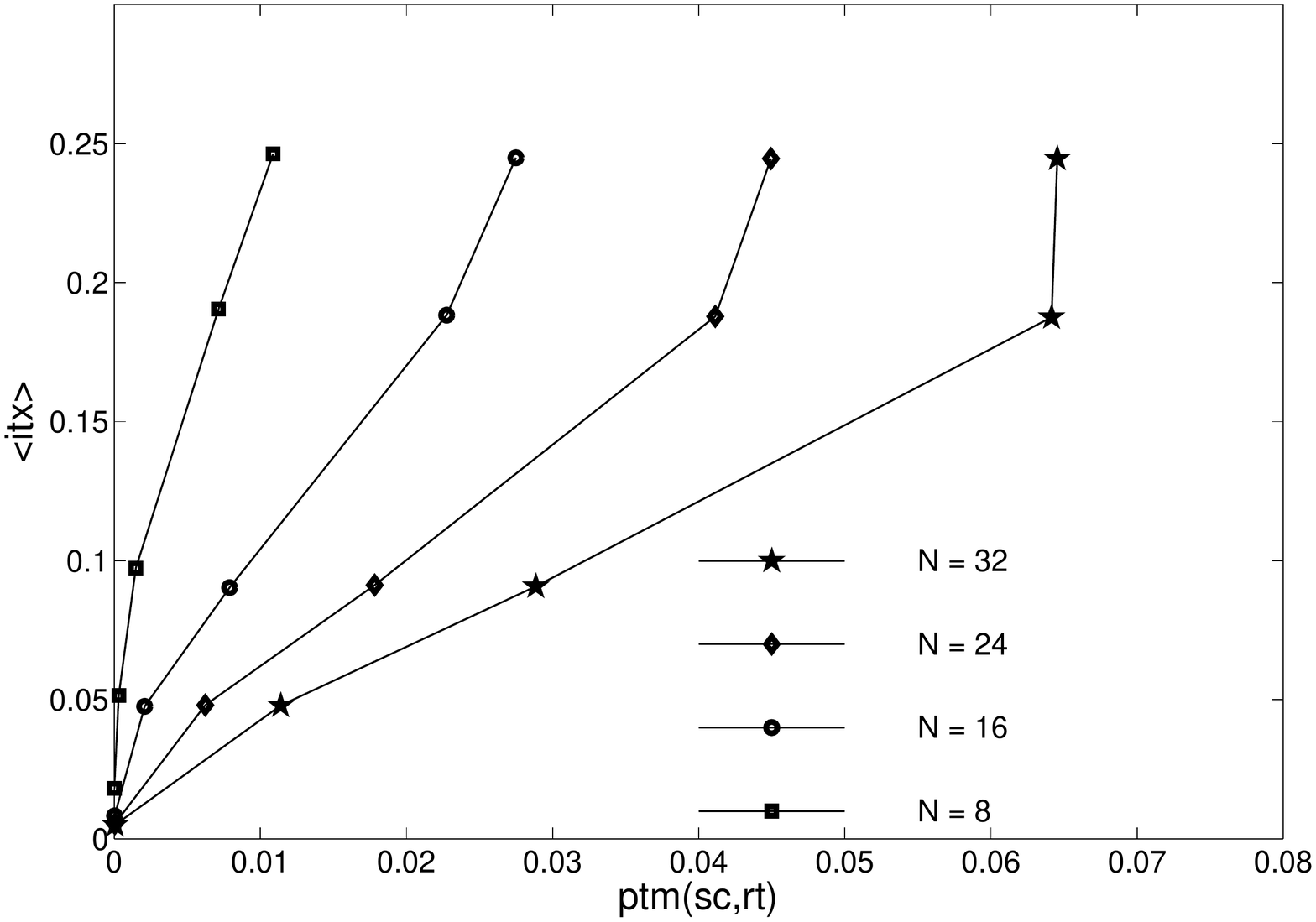}
\includegraphics[trim = 23mm 15mm 30mm 15mm,clip, width=6.0cm, height=5.8cm]{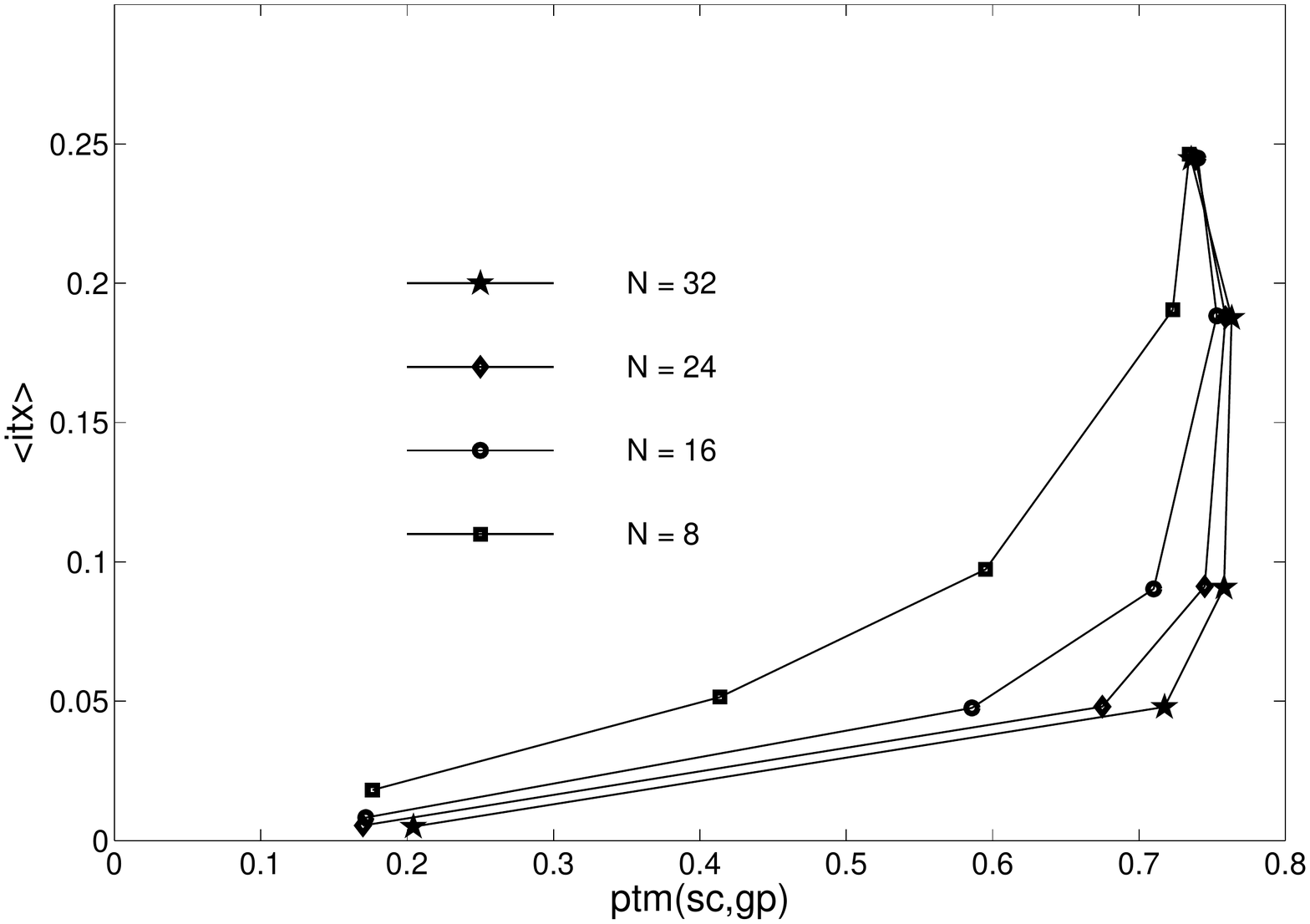}

\hspace{1.8cm}  (d) \hspace{3.6cm}  (e) \hspace{3.4cm}  (f)

\vspace{0.25cm}
 {\bf Fig. 2.} Intransitivity measure itx versus score $=\text{sc(k)}$, rating  $\text{rt(k)}$ and generalization performance $\text{gp(k)}$ (a--c) as scatter plots; relationships between the intransitivity measure itx,  kld, and ptm (d--f)
\end{figure}

\begin{figure}[tb]
\includegraphics[trim = 25mm 15mm 30mm 15mm,clip, width=6.0cm, height=5.8cm]{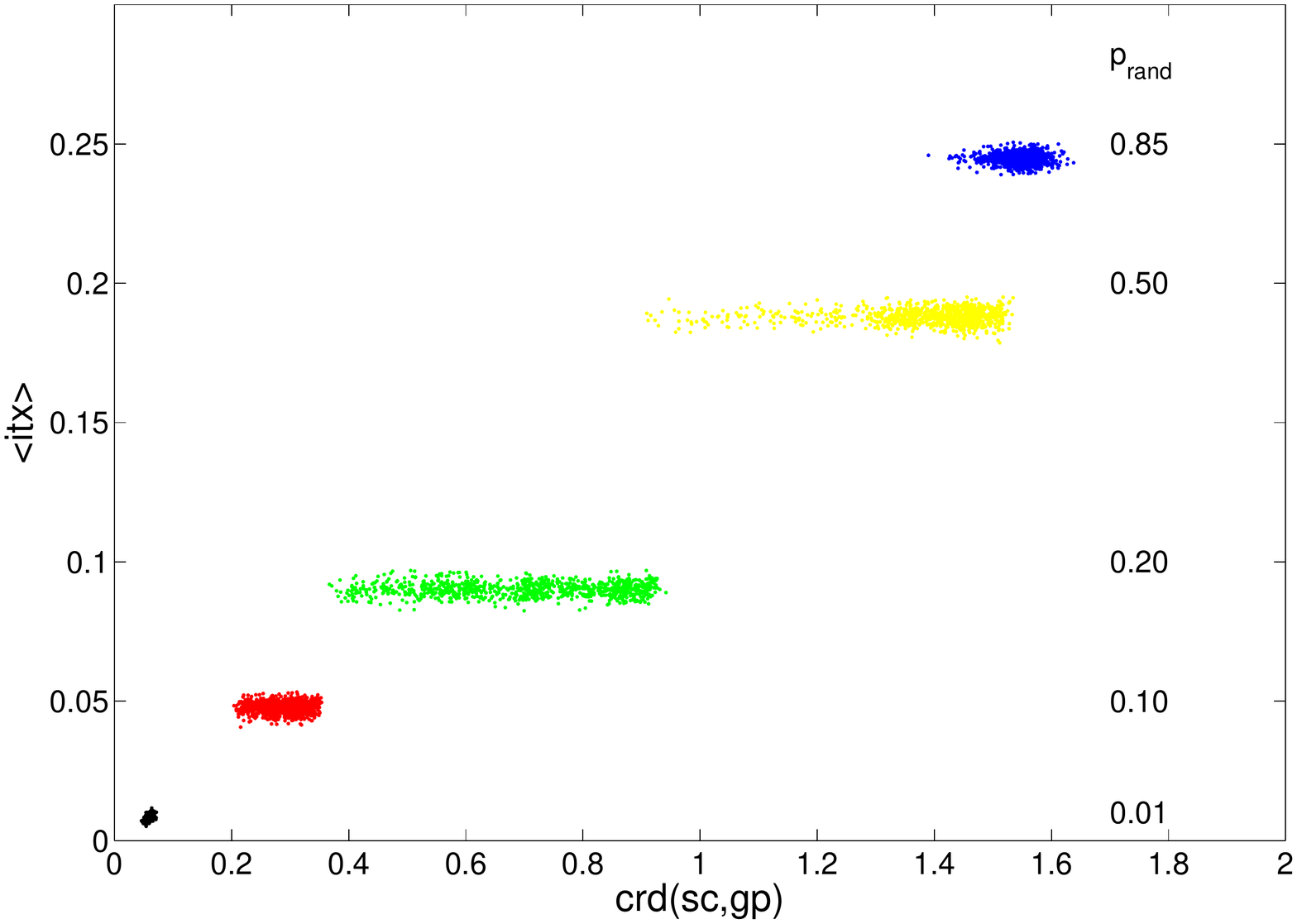}
\includegraphics[trim = 25mm 15mm 30mm 15mm,clip, width=6.0cm,height=5.8cm]{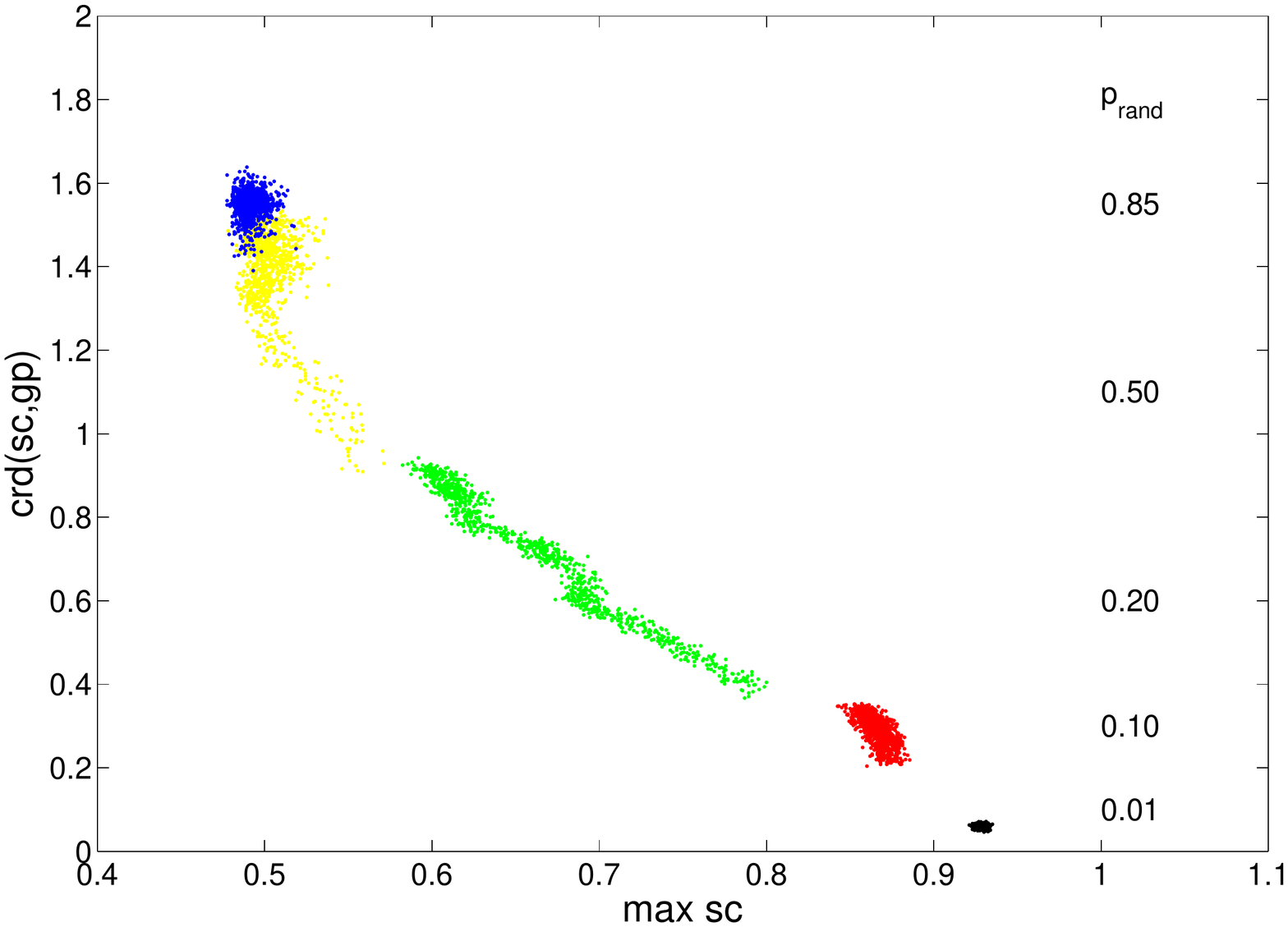}
\includegraphics[trim = 25mm 15mm 30mm 15mm,clip, width=6.0cm, height=5.8cm]{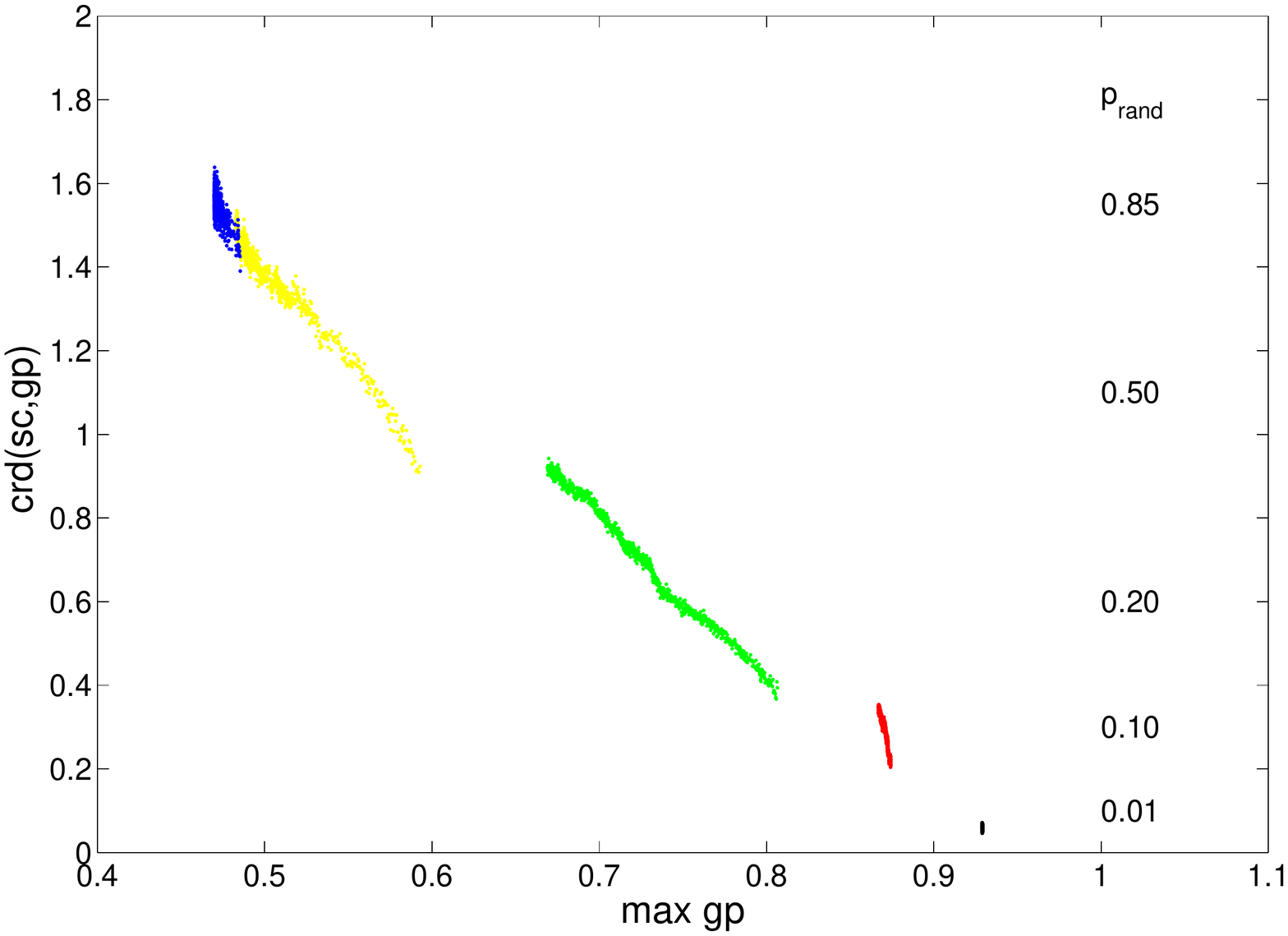}

\hspace{1.8cm}  (a) \hspace{3.6cm}  (b) \hspace{3.4cm}  (c)

\includegraphics[trim = 25mm 15mm 30mm 15mm,clip, width=6cm, height=5.8cm]{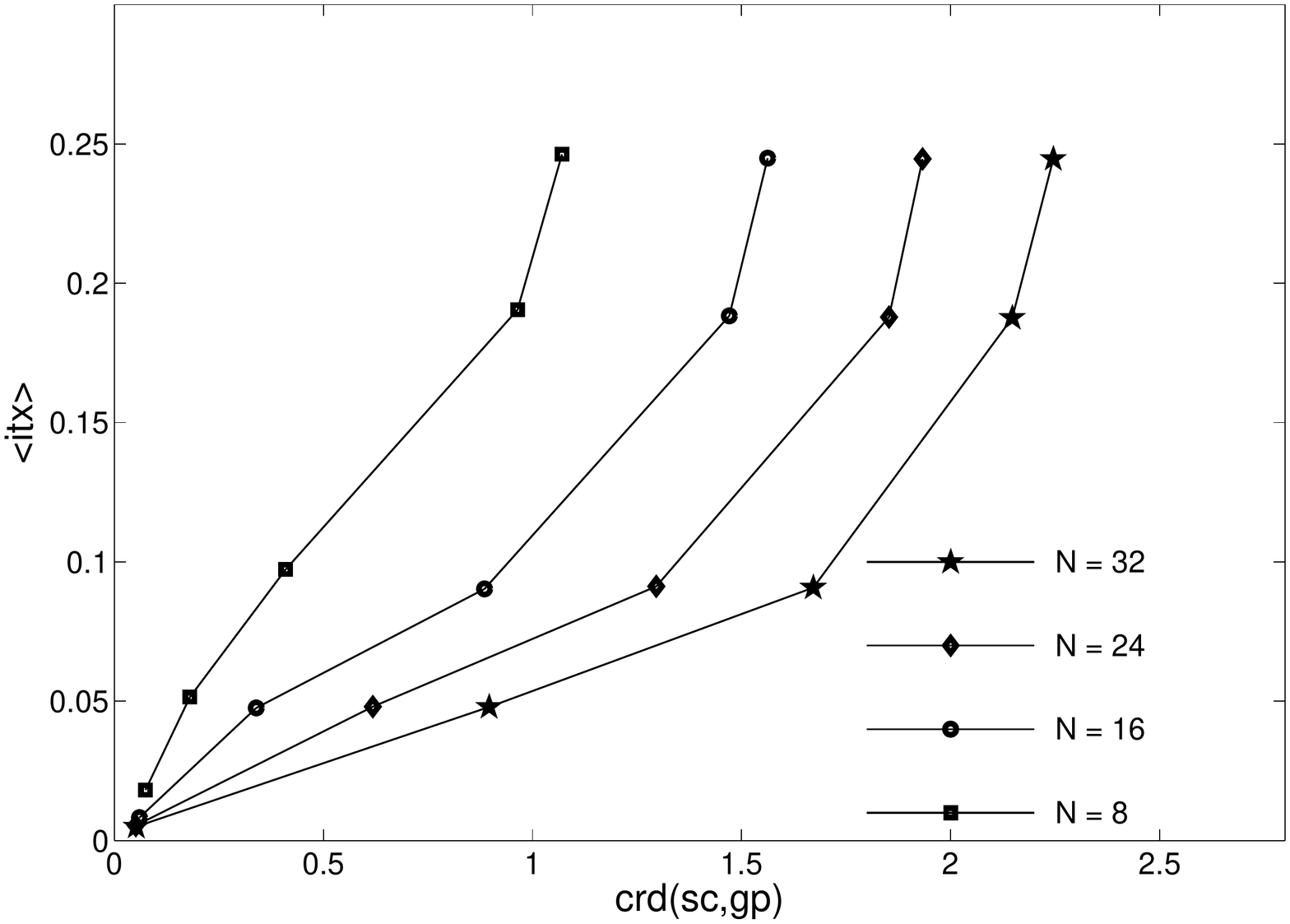}
\includegraphics[trim = 25mm 15mm 30mm 15mm,clip, width=6cm,height=5.8cm]{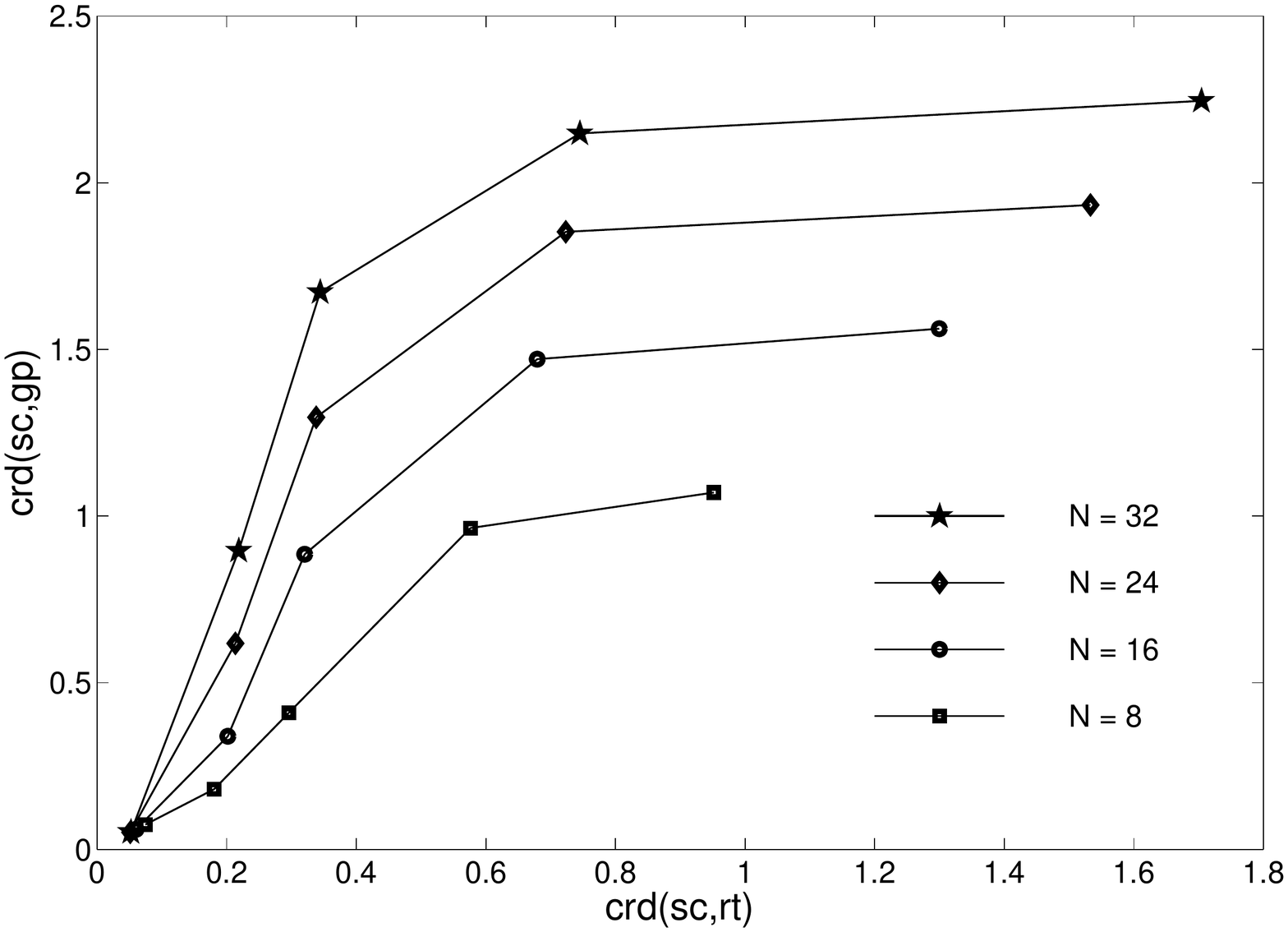}
\includegraphics[trim = 25mm 15mm 30mm 15mm,clip, width=6cm, height=5.8cm]{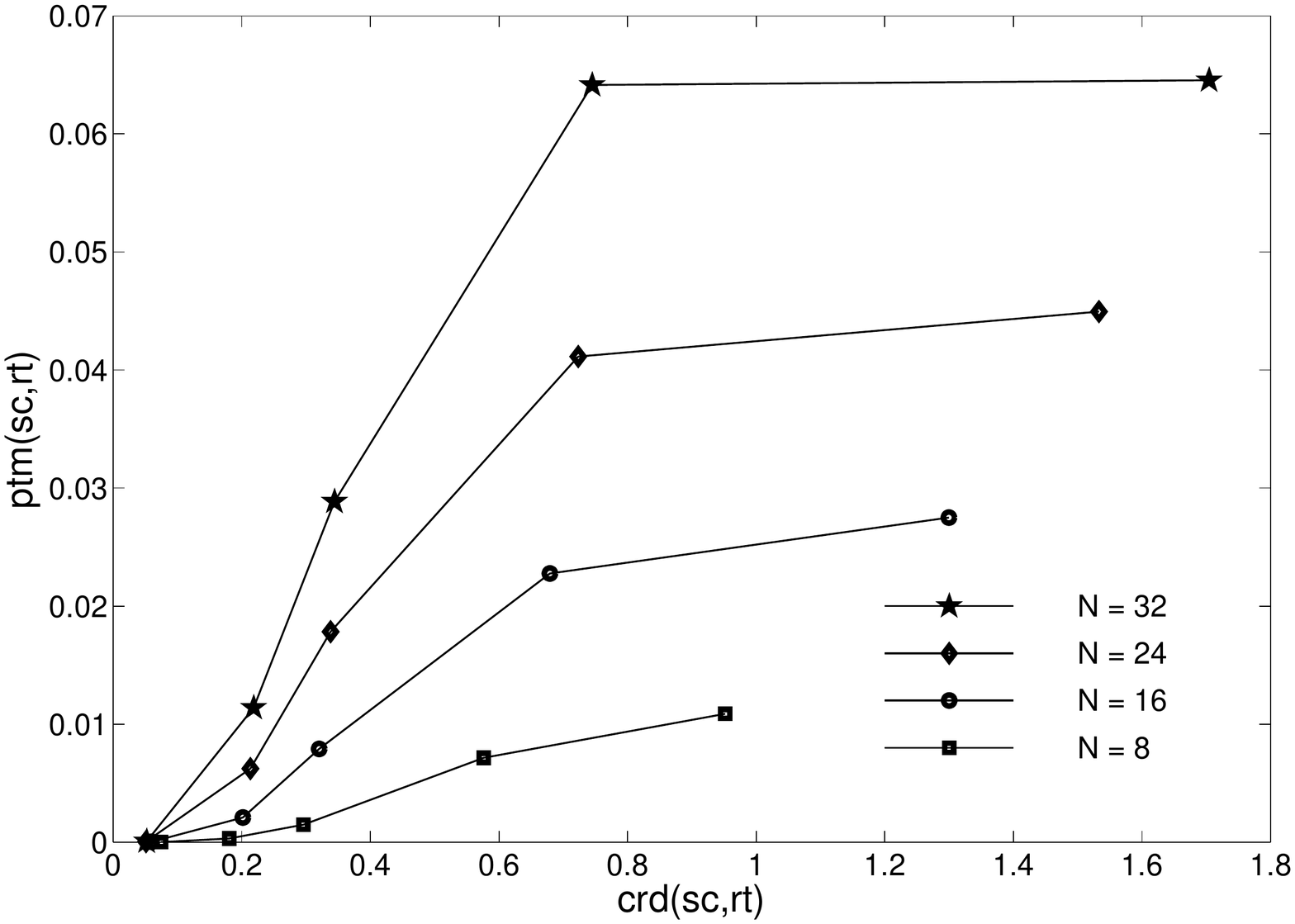}

\hspace{1.8cm}  (d) \hspace{3.6cm}  (e) \hspace{3.4cm}  (f)

\vspace{0.25cm}
 {\bf Fig. 3.}  Intransitivity measure crd versus time average itx, max sc, and max pg (a--c) as scatter plots; relationships between the intransitivity measure crd,  itx, and ptm (d--f)
\end{figure}

\section{Numerical experiments}
The report of experimental results starts with the time evolution of the simple random game introduced in the last section. Fig. 1 shows the scores, ratings and generalization performances for two different percentages of random results ($p_{rand}=0.01$ and  $p_{rand}=0.75$) and $N=6$ players and 20 instances of a round robin. The experiments are initialized with ratings $\text{rt}(0)$ being slightly spread around the value $\text{rt}(0)=1600$.  The randomly determined chance to win or to lose is evenly distributed.  The figures show curves for a single instance of the randomly determined part of the game outcomes. Hence, the curves are not meant to be statistically significant, but for illustrating typical behaviour only. 
 It can be seen that for a low number of random outcome (Fig. 1a,b,c)  the scores each player achieves are mainly defined by the initial rating. Small differences in the initial rating are amplified and lead to well--sorted long--term rankings of both the rating and generalization performance. For a large number of random game results (Fig. 1d,e,f) the scores are almost purely chance which is evenly distributed.  Consistently, ratings and  generalization performances tend to approach the expected value implied by the underlying distribution for all players alike. 
The next experiments address the relationships between static (game--induced) intransitivity expressed by the measure itx and quantities representing subjective as well as objective fitness. The scatter plots given in Fig. 2a,b,c are for $N=16$ players and five levels of $p_{rand}$. The experimental setup includes a repetition of each run for 100 times for 1000 instances of the round robin, where the first 200 instances are discarded to omit transients. Note that this gives a sufficient number of instance according to the bounds of generalization performance~\cite{chong08,chong12}. Hence, the results can be seen as statistically significant. In fact, the 99\% confidence intervals are so small that they are not depicted in the figures. It can be seen that for each level of randomness in the game, we obtain a distinct level of static intransitiviy measured by itx, where rising  $p_{rand}$ also increases itx. What is interesting is that there is almost no variation in score, rating or generalization performance for a given level of intransitivity itx. This indicates that static  intransitivity seems to have little influence on neither subjective nor objective fitness. This is particularly visible for low levels of $p_{rand}$ where these is a clear sorting according to the range of fitness a given player achieved but this is not connected to differences in the itx. A next experiment explores the relations between the index based measure itx and the probabilistically motivated measure kld, see Fig. 2d. Here as well as in the following figure the results are for four different number of players $N$ ($N=8,16,24,32$) and five levels of random results $p_{rand}$; the time--average itx, denoted $\langle \text{itx}\rangle$, is given, the quantities are normalized according to the number of players.  It can be seen that between both quantities there is a  proportional relationship, which on the one hand generally confirms the result in~\cite{samo13}, but does not show that itx is more brittle that kld. For the studied game it can be concluded that both quantities are interchangeable.  Next, the relationship between the static intransitivity and the dynamic intransitivity measure player--wise temporal mismatch (ptm) is studied, Fig. 2e,f. For the mismatch ptm based on rating as objective fitness (Fig. 2e) there is a linear relationship, at least for small values of $p_{rand}$, for the mismatch ptm based on generalization performance (Fig. 2f), no sensible conclusions about relations can be drawn. Finally, we focus on the rank--based dynamic intransitivity measure   collective ranking difference (crd), see Fig. 3. The Fig. 3a,b,c again shows scatter plots, now for the crd based on score as subjective fitness and generalization performance as objective fitness for $N = 16$ players and five levels of $p_{rand}$. It can be seen that although different $p_{rand}$ give different $\text{itx}$, there is no difference in the crd, Fig. 3a. However,  there is a almost linear relation  between the measure crd and  max sc and  max gp, at least for lower levels of $p_{rand}$. This can be interpreted as the crd scaling with the time evolution of the subjective and objective fitness, but not with the time evolution of static intansitivity, compare to Fig. 2a,b,c, which does not show such a characteristics. For the time--averages, there is  a scaling for different number of players and different levels of randomness $p_{rand}$, see Fig. 3d. In Fig. 3e, the relation between the crd based on rating as objective fitness (crd(sc,rt)) and the crd based on generalization performance as objective fitness (crd(sc,gp)) is shown.   It can be seen that both quantities scale piece-wise linear for $p_{rand}$ not very large, which allows to conclude that both quantities account for the same intransitivity properties. Finally, the relation between the crd and ptm is shown, Fig. 3f. It can be seen that the ptm scales weaker than the crd, particularly for a small number of players and high randomness, and it can be conjectured that the crd is a more  meaningful coevolutionary intransitivity measure than the ptm.

\section{Conclusions}
This paper is a contribution to the ongoing discussion about the effect of intransitivities on coevolutionary progress.  An approach was presented that allowed to link   a rating-- and ranking--based measuring  approach of intransitivity~\cite{samo13}  with a framework of  fitness landscapes to enable  analyzing the relationship between objective and subjective fitness. For experimentally illustrating the approach a simple random game with  continuously tunable  degree of randomness was proposed. Apart from the random, the game results depend on the ratings of the players, which reflect the past success of each player. Thus, the game proposed characterizes many real--world games as their outcome is also a function of chance as well as of predictions based on game history.  

For studying the effect of intransitivity, measures were explored. In extension  of existing static intransitivity measures, dynamic measures that can account for coevolutionary intransitivity were proposed. These measure are based on rankings between subjective and objective fitness and it was shown that
coevolutionary intransitivity can be understood as a ranking problem, and hence be accounted for by ranking statistics. To enlarge the scope of the presented approach, as a next step the intransitivity measures could be studied for other types of game for instance social games as the  iterated prisoner's dilemma  or board game such as Othello.

\end{document}